\documentclass[letterpaper, 10 pt, conference]{ieeeconf}  %

\IEEEoverridecommandlockouts                              %

\usepackage{graphics} 
\usepackage{amsmath} %
\usepackage{amssymb}  %
\usepackage{subcaption}
\usepackage{multicol}
\usepackage{mathtools}
\usepackage[dvipsnames]{xcolor}
\usepackage[most]{tcolorbox}
\usepackage{eso-pic}
\usepackage{tikz}
\usetikzlibrary{matrix, fit, backgrounds}

\newcommand{\hl}[2]{
	\begin{scope}[on background layer]
		\node [fit={#1}, fill=#2,inner sep=-1pt] {};
\end{scope}}

\newtcolorbox{mytextbox}[1][]{%
	sharp corners,
	enhanced,
	colback=white,
	height=6cm,
	attach title to upper,
	#1
}
\newcommand{\vect}[1]{\boldsymbol{#1}}

\title{\LARGE \bf
HEDRA: A Bio-Inspired Modular Tensegrity Soft Robot With Polyhedral Parallel Modules 
}

\author{Vishal Ramadoss*, Keerthi Sagar*, Mohamed Sadiq Ikbal, Jesus Hiram Lugo Calles and Matteo Zoppi %
	\thanks{* Equal contribution. V. Ramadoss, K. Sagar,  M.S. Ikbal, J.H. Lugo Calles and M. Zoppi are with PMAR Robotics Group, Department of Mechanical, Energy, Management and Transportation Engineering, University of Genoa, Genoa (GE), 16145, Italy. Email: 
		{\tt\small \{keerthi, lugocalles, zoppi\}@dimec.unige.it, \{vishal.ramadoss, mohamedsadiq.ikbal\}@edu.unige.it} }}

\begin{document}

\maketitle
\thispagestyle{empty}
\pagestyle{empty}

\begin{abstract}
There is a surge of research interest in the field of tensegrity robotics. Robots developed under this paradigm provide many advantages and have distinguishing features in terms of structural compliance, dexterity, safety, and weight reduction. This paper proposes a new robotic mechanism based on tensegrity (`tension-integrity') robots and reconfigurable modular robots. The specific actuation schemes for this tensegrity robot with multiple degrees of freedom are presented. This article describes an easy-to-assemble 350 mm tensegrity based robot prototype by stacking a series of rigid struts linked with tensegrity joints that have no direct rigid contact with each other.  The functionality of the proposed robot is validated by the experimental results by integrating the polyhedral parallel structure as its skeleton and series of tensegrity joints. The proposed manipulator is capable of reaching bending angles up to 76 degrees. An adaptive cable driven underactuated robotic gripper is designed and attached to the tensegrity manipulator for grasping objects in different shapes, weights, and sizes.
 
\end{abstract}

\section{INTRODUCTION}
Traditional rigid robots are fast and precise systems that are used in manufacturing environments. The advent of collaborative robots have significantly reduced the costs and have the ability to work safely alongside humans. This has led to the evolution towards human friendly robotics. In parallel, soft robotics have garnered widespread attention and offers solutions that cannot be addressed by robots built from rigid bodies. An alternative design methodology to either rigid or soft robots are tension-integrity systems. 

The emerging field of tensegrity robotics aims to combine the advantages of conventionally stiff and the continuous compliant mechanisms. This results in a smaller form factor with regards to the footprint of the robot and lower power requirements. Novel applications are being proposed for robotic systems such as robots for space explorations \cite{post2018proof}, search and rescue and rehabilitation. Such robots are pioneering soft robotic concepts which combine biomimetics and tensegrity frameworks. They are lightweight \cite{bruce2014design}, reconfigurable \cite{belke2017mori} and compliant \cite{friesen2018tensegrity} similar to soft robots and performs complex motions. Tensegrity demonstrates the natural balance of forces where the compression members are suspended in the dynamic tension network.

Significant work has been done to leverage the concepts of tensegrity in robotic systems, which transforms their shape thereby performing crawl \cite{tietz2013tetraspine} \cite{paul2006design}, swim \cite{chen2019swimming} \cite{shintake2020bio}, roll \cite{chen2017soft}, climb \cite{friesen2014ductt} and fly \cite{moored2006optimization} \cite{zha2020collision}. The contribution in the field of biotensegrity \cite{scarr2014biotensegrity}\cite{flemons2012bones} produced manipulators that emulate the human elbow \cite{lessard2016lightweight}, arm \cite{li2020new} \cite{lessard2016bio}, shoulder \cite{levin1997putting}\cite{baltaxe2016simulating} and vertebrae \cite{sabelhaus2015mechanism} \cite{sabelhaus2018design}\cite{9050924}.
  \begin{figure}[t]
  	\centering
  	\includegraphics[width=0.65\linewidth,keepaspectratio]{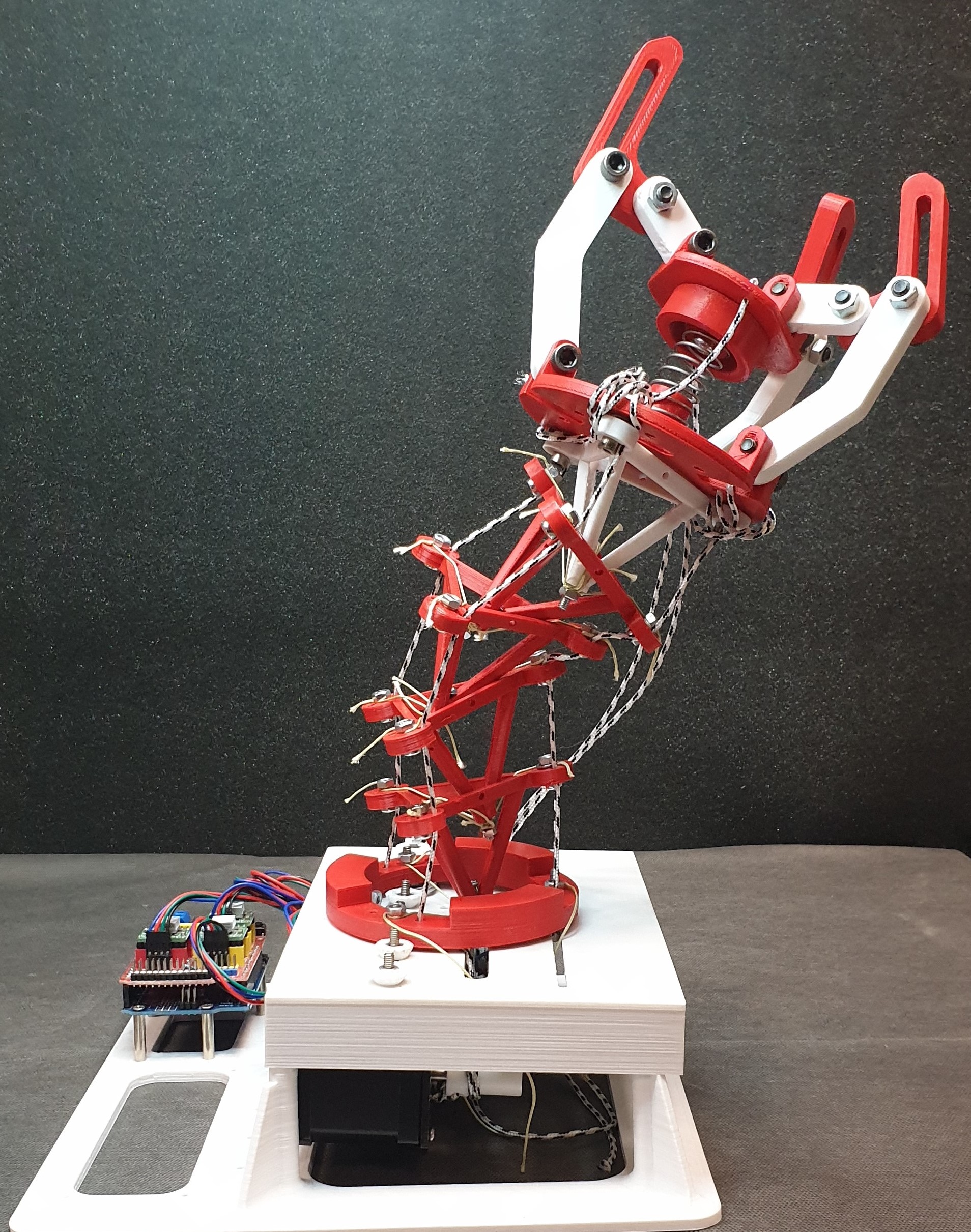}
  	\caption{Hedra at stowed configuration}
  	\label{fig:hedraneutral}
  \end{figure}
\begin{figure*}[t]
	\centering
	\includegraphics[width=0.75\textwidth,keepaspectratio]{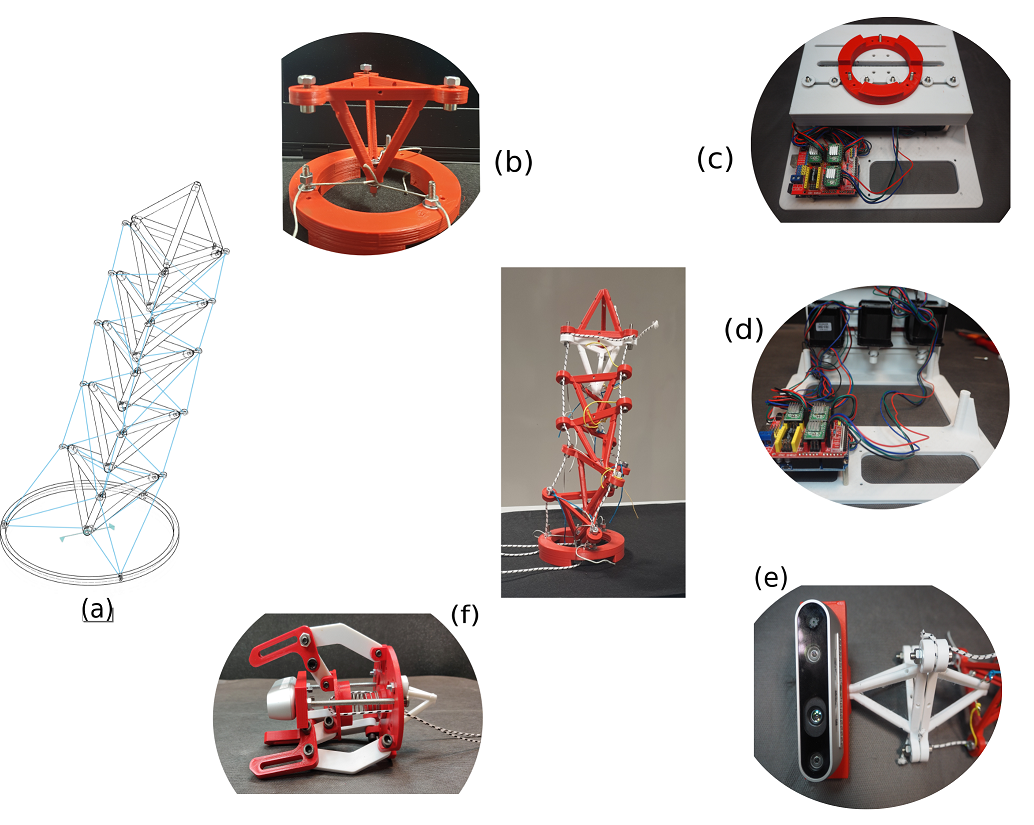}
	\caption{(a) Wireframe of tensegrity manipulator with tetrahedral parallel modules with 3-Dof tensegrity joints (b) The structure of tensegrity joint with axial and horizontal tension network. The cable outlets are fixed by the bolts and nuts. (c) base plate with barrel jack connectors equipped with motor drivers powering the tensegrity module (d) closer view of the interface between motors and motor driver (e) closer view of the octahedral geometry end-effector mounted with Intel Realsense D435i camera (f) Adaptive cable-driven gripper module  }
	\label{fig:hedramain}
\end{figure*}
Existing frameworks  results in configuration with a large number of bars and cables making fabrication and assembly challenging with constraints on space for actuators and sensors. Owing to this fact, most of the `bar-only' tensegrities are tested in simulation or workbenches. There is a need for an archetype to bridge the gap between simulation and hardware, that allows innovative bio-inspired geometries rather than limiting the forms of tensegrity. Bioinspired vertebrae morphologies are utilized and its geometric compliance is exploited to simplify the design process. This is the motivation behind the development of our proposed bio-inspired modular tensegrity soft robot with polyhedral parallel modules (HEDRA) as shown in Figure \ref{fig:hedraneutral}.

The contributions of this work can be summarized as follows:
\begin{itemize} 
	\item A low-cost, lightweight, easy to assemble modular robot which have distinct softness with granular rigidity for education and research;
	\item An alternative to conventional rigid robotic mechanisms with advantages of omnidirectional compliance and variable stiffness of redundant serial-parallel robots;
	\item Minimal design changes while stacking modules and addition of parallel stages with ability to build into closed-loop and tree robots;
	\item Reconfigurable end-effector with vision based system and adaptive cable-driven gripper module;
\end{itemize}

In what follows, we first describe the concept design and fabrication, including the overall geometry of the proposed robot in Section II. A simple modeling framework is described in Section III. The simulations and proof-of-concept is elucidated in Section IV. The concluding remarks are drawn in Section V.  

\section{MATERIALS AND METHODS}
\subsection{Concept design and Fabrication}
Fig. \ref{fig:hedraneutral} shows the overall structure of the bio-tensegrity robot in stowed configuration. It is composed of five tetrahedral rigid platforms which are linked with vertical tensile cables and one tetrahedron at the top serving as end-effector equipped with tool mount. Structural compliance is achieved due to the elastic property of tensile components.

\begin{figure}[!htb]
	\centering
	\includegraphics[width=0.75\linewidth,keepaspectratio]{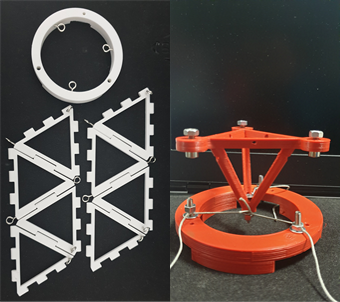}
	\caption{a) Easy to assemble modular prototype with eyelets b) Tensegrity joint with 3 horizontal saddle cables and viapoints for active cables fixed by bolts and nuts}
	\label{fig:hedraone}
\end{figure}

\begin{figure}[!t]
	\centering
	\includegraphics[width=0.5\linewidth,keepaspectratio]{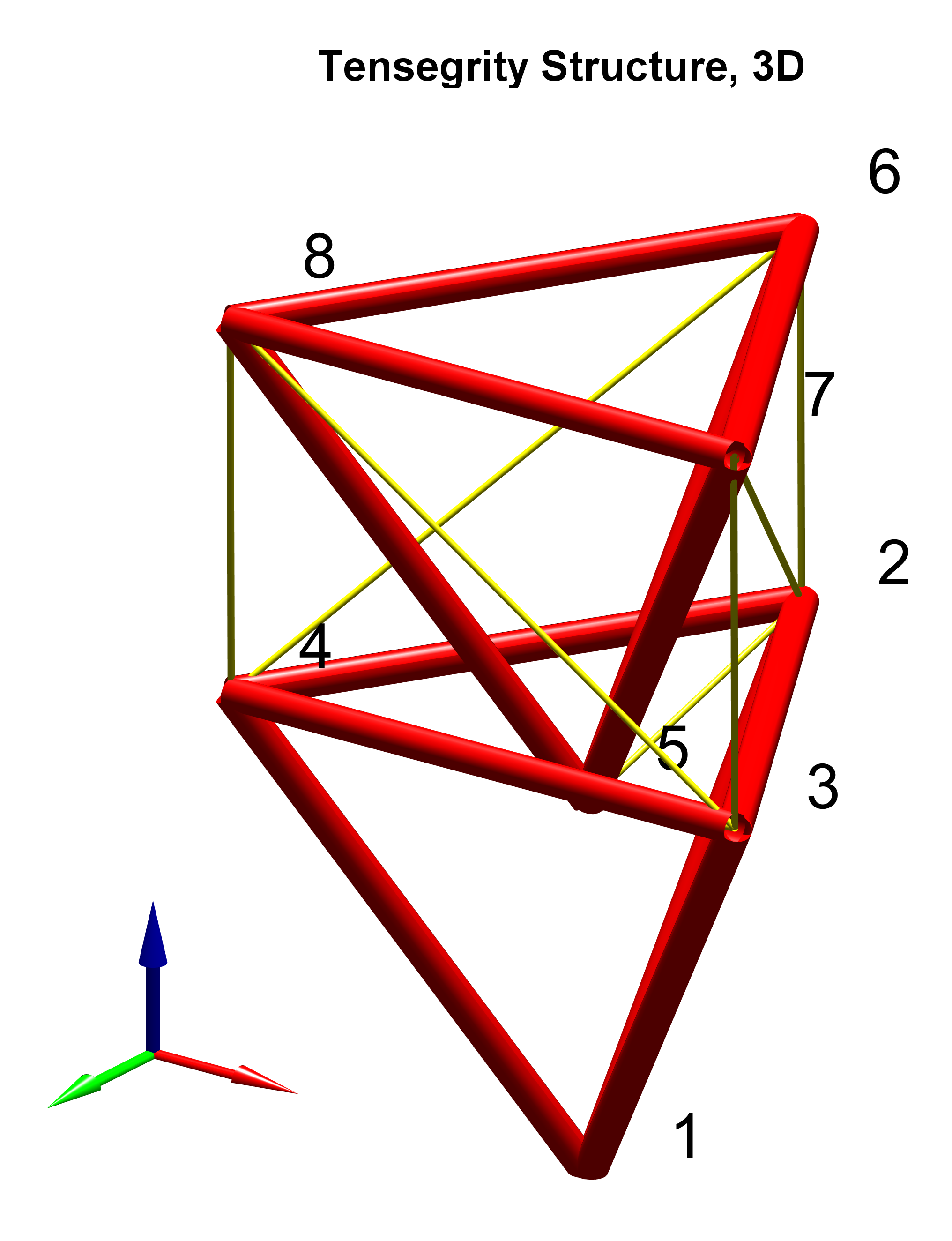}
	\caption{A three-dimensional, tetrahedral tensegrity manipulator. Black dents are nodes, red edges are bars (compressive), and yellow edges are cables (tensile). x-,y-,z- axis is denoted by red, green and blue color respectively.}
	\label{fig:nodes}
\end{figure}

The tetrahedron acts as the compression element of the tensegrity joint. The tensile elements are the cables that are divided into two groups, the horizontal saddle cables and the axial cables as shown in Fig. \ref{fig:nodes}. The cable tension forces acting axially serves to pull the stacked tetrahedra together. The horizontal tension network is connected from the bottom node of the top tetrahedra to support the bottom tetrahedra in suspension. The top tetrahedra is under downward tension due to the axial tension network and also the saddle cables can serve the upward tension and resist the downward forces. This resistance from the saddle cables is augmented to the overall network of tensegrity joints and the load is distributed. This is the primary reason behind the motion of the top tetrahedron around the bottom tetrahedron without direct contact thereby acting as tensegrity joint. Then, the tensegrity joints are connected in series to have a stacked manipulator. The joint has 3 degrees of freedom having a virtual ball joint center at the center of saddle cable network as shown in Fig. \ref{fig:hedraone}(b).
In order to transform the redundant series-parallel mechanisms into a compliant mechanism, bioinspired backbone \cite{sabelhaus2018design} and lamprey \cite{tytell2010interactions} morphologies are utilized.

The robot core entity, the tetrahedron and the base are 3D printed in PLA, diameter 1.75 mm, using Delta Wasp 3D printer. The tetrahedron unit weighs 14g and an octahedron topology is chosen as end-effector for mounting the Intel Realsense D435i camera (260 gms). A three finger linkage based gripper mechanism weighing 280 gms is developed separately to perform grasping. The prototype is driven by three active cables made of Dyneema ultra-high molecular weight polyethylene (UHMWP) wires with diameter of 2mm and Young's Modulus of 110 GPa. The saddle cables are taut using Spectra Extreme Braid fishing line with diameter of 0.68 mm and with load capacity of 60 kgs. NEMA 17 stepper motors with A4988 motor drivers were used for the initial stages of testing in the laboratory setup to visualize the bending and axial twist motion. Brushed DC motors (Pololu 156:1 20D x 44L ) with magnetic encoder 20 counts per revolution with pulley systems, MC33926 motor driver and a controller board that offers on-board sensory measurements and communication was utilized to control the active cables. 

\begin{figure}[ht]
	\centering
	\includegraphics[width=0.65\linewidth,keepaspectratio]{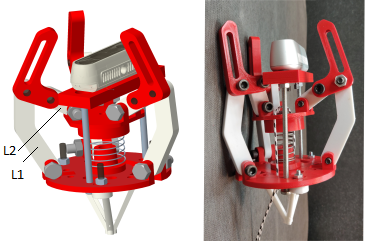}
	\caption{Left: CAD rendering of the adaptive cable driven gripper equipped with Intel Realsense Camera. Right: 3D printed prototype}
	\label{fig:rs}
\end{figure}

 Different cable routing schemes have been proposed previously for multi link cable driven robots \cite{thesisvish} \cite{asmevishal} \cite{vishaluar}. We initially model with minimally actuated cable setup with 3 active cables and investigate the clustered actuation model with 6 active cables based on the previous works on actuation strategy. For experiments, minimally actuated Hedra is utilized. The unit tetrahedron has edge length of 70 mm, the height from the base of the tetrahedron is 51.48 mm, the thickness of each edge is 3 mm. The scaled version of tetrahedron is used for prototyping from simulation. The three-jaw gripper, shown in Fig. \ref{fig:rs} is constructed using closed loop linkage mechanisms with link1 (L1) length as 72.45 mm, link2 (L2) as 23mm, ground link length is 54 mm and the distance between the base plate and the base of realsense gripper. The initial length of spring is 35mm and the compressed length is 15 mm respectively.
\\

\section{Topological Notation}
A tensegrity structure is defined by a graph $G= (V,E)$, with nodes $V$ and edges  $E$. There are $n$ nodes, where each $\mathbf{a}_{i} \in V $, $i=1 \dots n$ represents a point in $\mathbb{E}^{3}$ where the structural members connect. There are $m$ edges, where each edge $e_{k} \in E$, $k=1 \dots m$ represents a structural member that connects two nodes. Forces $P= \lbrace{\vect{p}_{1},\dots,\vect{p}_{n}}\rbrace,\vect{p}_{i} \in \mathbb{E}^{3}$ are applied at each node. Let $\vect{x}, \vect{y}, \vect{z} (\in \mathbb{R}^{n})$ denote the nodal coordinate vectors in the $x,y$ and $z$ directions. The structural members consists of $r$ bars that can take compression or tension loading, or $s$ cables, that can only take tension loading. The connectivity information of members can be expressed by the branch-node or incidence matrix, $\mathbf{C}\in (\mathbb{R}^{(s+r)\times n})$. If member $i \in \lbrace{1,\dots,(s+r)}\rbrace$ connects nodes $k$ and $j$ ($k<j$), then the $k$-th and $j$-th columns in $\mathbf{C}$ are set to 1 and -1 respectively, as
\quad
\begin{equation}
 \mathbf{C}^{(u,v)} =
\begin{cases}
1 & \text{if $u=i$ , $v=k$}\\
-1 & \text{if $u=i$ , $v=j$}\\
0   & \text{if else}
\end{cases}
\end{equation}

For the tensegrity structure shown in Fig. \ref{fig:nodes}, the connectivity matrix is given by

\[\mathbf{C}_{s}=
\begin{tikzpicture}[scale=0.95, every node/.style={transform shape},baseline=-0.5ex]

\matrix [matrix of math nodes,every cell/.style={scale=0.75},left delimiter={[},right delimiter={]}] (m)
{
	0  &   1 &    0  &   0  &  -1 &    0  &   0  &   0 \\
	0  &   0 &    1  &   0  &  -1 &    0  &   0  &   0 \\
	0  &   0 &    0  &   1  &  -1 &    0  &   0  &   0 \\
	0  &   1 &    0  &   0  &   0 &   -1  &   0  &   0 \\
	0  &   1 &    0  &   0  &   0 &    0  &  -1  &   0 \\
	0  &   0 &    1  &   0  &   0 &    0  &  -1  &   0 \\
	0  &   0 &    1  &   0  &   0 &    0  &   0  &  -1 \\
	0  &   0 &    0  &   1  &   0 &    0  &   0  &  -1 \\
	0  &   0 &    0  &   1  &   0 &   -1  &   0  &   0 \\
};  
\hl{(m-1-1) (m-9-4)}{blue!40, opacity=0.1,rounded corners} 
\hl{(m-1-5) (m-9-8)}{blue!40, opacity=0.5,rounded corners} 

\end{tikzpicture}\]

\[\mathbf{C}_{r}=
\begin{tikzpicture}[scale=0.95, every node/.style={transform shape},baseline=-0.5ex]

\matrix [matrix of math nodes,every cell/.style={scale=0.75},left delimiter={[},right delimiter={]}] (m)
{
1 &   -1  &   0  &   0  &   0  &   0  &   0  &   0 \\
1 &    0  &  -1  &   0  &   0  &   0  &   0  &   0 \\
1 &    0  &   0  &  -1  &   0  &   0  &   0  &   0 \\
0 &    1  &  -1  &   0  &   0  &   0  &   0  &   0 \\
0 &    1  &   0  &  -1  &   0  &   0  &   0  &   0 \\
0 &    0  &   1  &  -1  &   0  &   0  &   0  &   0 \\
0 &   0   &  0   &  0   &  1  &  -1   &  0   &  0 \\
0 &    0  &   0  &   0  &   1  &   0  &  -1  &   0 \\
0 &    0  &   0  &   0  &   1  &   0  &   0  &  -1 \\
0 &    0  &   0  &   0  &   0  &   1  &  -1  &   0 \\
0 &    0  &   0  &   0  &   0  &   1  &   0  &  -1 \\
0 &    0  &   0  &   0  &   0  &   0  &   1  &  -1 \\
};  
\hl{(m-1-1) (m-12-4)}{red!40, opacity=0.1,rounded corners}                                                         
\hl{(m-1-5) (m-12-8)}{red!40, opacity=0.5,rounded corners} 

\end{tikzpicture}\]

\begin{equation}
\mathbf{C}=\begin{bmatrix}
\mathbf{C}_{s}\\
\mathbf{C}_{r}
\end{bmatrix}
\end{equation}

where $\mathbf{C}_{s}$ (indicated in blue) represents the cable connectivity with $s=9$ cables and $\mathbf{C}_{r}$ (indicated in red) represents the bar connectivity with $r=12$ bars and the size of $\mathbf{C}$ for the robot in Fig. \ref{fig:nodes} is $21 \times 8$. The lighter color represent the nodes for the bottom tetrahedron which is fixed to the ground and the dark-shaded rows represent the moving tetrahedra. 

The biotensegrity robot, in general has $n$ tetrahedrons, in three dimensions. The radius of the tetrahedron is $r$ and the origin of the coordinate system is at the common point of the lowermost tetrahedron, where the flat polygon base and the triangular faces are connected. The initial coordinates of the base tetrahedron are parameterized as

\begin{equation}
\mathbf{S}_{i}=
\begin{bmatrix}
r\sin(0)&       r\cos(0)&             h\\
r\sin(\alpha)&  r\cos(\alpha)&        h\\
r\sin(-\alpha)& r\cos(-\alpha)&       h\\
\end{bmatrix}
\end{equation}

where the rows are the cartesian coordinates of each tetrahedron in space, $r$ is radius of a circle that is inscribed by the top triangle of the tetrahedra, $h$ is the height of the tetrahedra and $\alpha= 2\pi/3$. The state space matrix of each tetrahedron is defined by the three position variables due to translation and three orientation variables. After transformation of the tetrahedron, the location of points in space is represented by

\begin{equation}
\mathbf{S}^{'}_{i}=\mathbf{S}_{i}R_{\theta_{i},\gamma_{i},\phi_{i}}+UT_{x_{i},y_{i},z_{i}}
\end{equation}
where $R$ is the rotation matrix for each Euler angle, $U$ is the unitary matrix and $T$ is translation matrix.
\section{Control and Simulation}
\subsection{Force Density Method For Inverse Kinematic Control Policy}
The force density method (FDM) for solving cable network problems was described in \cite{schek1974force} \cite{tran2010advanced} \cite{tibert2003review} \cite{gan2020computational}. The method used was presented in works on related topologies of tensegrity robot \cite{friesen2014ductt} \cite{sabelhaus2015mechanism}. For the sake of clarity, the static equilibirum conditions for the structure is briefly summarized.  The equilibrium equations of forces at each node of the cable network is utilized to solve the inverse kinematics of the proposed tensegrity robot, with the imposed assumptions being:
\begin{itemize} 
	\item The nodes are fixed in space (known coordinates) and do not translate.
	\item All forces are known and are only exerted at nodes.
	\item The connections between the members are friction-less ball joints.
	\item The structural members are perfectly rigid, do not deform and do not change length.
\end{itemize}
The force balance condition in three dimensions for static equilibrium of the structure can be stated as

\begin{equation}
\begin{aligned}
\mathbf{C}^{\mathsf{T}}\text{diag}(\vect{q})\mathbf{C}\vect{x} =\vect{p}_{x} \\
\mathbf{C}^{\mathsf{T}}\text{diag}(\vect{q})\mathbf{C}\vect{y} =\vect{p}_{y}\\
\mathbf{C}^{\mathsf{T}}\text{diag}(\vect{q})\mathbf{C}\vect{z} =\vect{p}_{z} 
\end{aligned}
\end{equation}
\quad

such that  $\vect{x},\vect{y},\vect{z}$ represents the nodal coordinate vectors, $\vect{p}_{x},\vect{p}_{y},\vect{p}_{z}$ represents the external loads applied to the nodes in the $x, y$ and $z $ directions, $\vect{q}= \lbrace{q_{1},q_{2},\dots,q_{s+r}}\rbrace^\mathsf{T}$ is the force density vector, where $q_{i}$ is the ratio between the force, $f_{i}$, and the length, $l_{i}$, such that 

\begin{equation}
q_{i}=\frac{f_{i}}{l_{i}}
\end{equation}
\quad

The equations are reorganized in the form 

\begin{equation}
\mathbf{A}\vect{q}=\vect{p},
\label{eqn:eqbm}
\end{equation}
where

\begin{equation}
\mathbf{A}=
\begin{bmatrix}
\mathbf{C}^{\mathsf{T}}\text{diag}(\mathbf{C}\vect{x})  \\
\mathbf{C}^{\mathsf{T}}\text{diag}(\mathbf{C}\vect{y}) \\
\mathbf{C}^{\mathsf{T}}\text{diag}(\mathbf{C}\vect{z}) 
\end{bmatrix}
\end{equation}
\quad
\begin{equation}
\vect{p}=
\begin{bmatrix}
\vect{p}_{x} \\
\vect{p}_{y}\\
\vect{p}_{z} 
\end{bmatrix}
\end{equation}
\quad
Equation (\ref{eqn:eqbm}) can be solved for $\vect{q}$ using the Moore-Penrose pseudoinverse with the general solution as

\begin{equation}
\vect{q}=\mathbf{A}^{+}\vect{p}+(\mathbf{I}-\mathbf{A}^{+}\mathbf{A})\vect{w},
\label{eqn:soln}
\end{equation}
\quad
where $(.)^{+}$ denotes the pseudoinverse of a matrix, $\mathbf{I}$ is the identity matrix, and $\vect{w}$ is arbitrary vector in $\mathbb{R}^{3n}$. The equations can be further split between the first $s$ rows and the last $r$ rows to denote the elements contributing to cable force densities and bar force densities such as 

\begin{equation}
\begin{bmatrix}
\vect{q}_{s} \\
\vect{q}_{r}
\end{bmatrix}=
\begin{bmatrix}
(\mathbf{A}^{+})_{s}\\
(\mathbf{A}^{+})_{r}
\end{bmatrix}\vect{p}
+(\begin{bmatrix}
\mathbf{I}_{s} &  0 \\
0  &  \mathbf{I}_{r} 
\end{bmatrix}- 
\begin{bmatrix}
(\mathbf{A}^{+}\mathbf{A})_{s}\\
(\mathbf{A}^{+}\mathbf{A})_{r}
\end{bmatrix})\vect{w},
\label{eqn:soln1}
\end{equation}
\quad

 An optimal set of cable forces can be found using quadratic programming approach that can be solved in bounded time and can be used for real time control. The optimization problem with an aim to minimize the amount of potential energy in the cables with the unilateral constraint of cables can then be implemented. After obtaining the solution for $\vect{q}$, the rest lengths can be calculated (for example, in the linear elastic case $f_{i}=K_{i}(l_{i}-l_{0i})$): 

\begin{equation}
l_{0i}=l_{i}(1-\frac{q_{i}}{K_{i}})
\label{eqn:restlength}
\end{equation}

\begin{figure}[ht]
	\centering
	\includegraphics[width=0.9\linewidth,keepaspectratio]{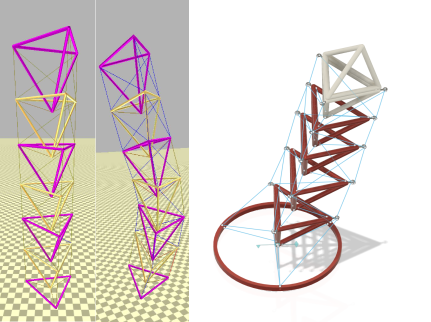}
	\caption{Left: Upright unactuated Hedra in equilibrium. Middle: Lateral Bending in Hedra with actuation in NTRT simulator. Right: CAD rendering of 5-link tensegrity manipulator where the blue lines are cable elements and the elements in red color are bars with end-effector bars in white color}
	\label{fig:rsa}
\end{figure} 

\subsection{NTRT Simulation}
\begin{figure}[t]%
	\centering
	\includegraphics[width=0.45\textwidth]{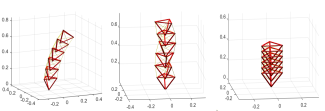}
	\caption{Inverse kinematic control policy simulation using force density approach. The working phases of the compliant manipulator. From left: bend, axial twist and contracted configurations}
	\label{fig:ma}
\end{figure}

The NASA Tensegrity Robotics Toolkit (NTRT) \cite{nasa} is an open-source library developed by the Intelligent Robotics Group for modeling and simulation of tensegrity robots based on the Bullet Physics engine. Kinematics, dynamics simulation and experimentation of previous works have been validated using NTRT. We have also utilized the features of NTRT simulator that has a custom linear cable model using Hooke's law forces and built the Hedra robot. The aim of the simulation was to check the rest lengths determined by the inverse kinematic algorithm positioning of the robot in a desired configuration in space as shown in Fig. \ref{fig:rsa}. This was further validated using simulations performed in MATLAB interface as shown in Fig. \ref{fig:ma}. 

The simulation was performed considering the radius of top tetrahedron as 0.1 m, height of tetrahedron as 0.15 m, the outer rim cable stiffness as 10000 N/m with minimum force density of 500 N/m.

\begin{figure}[t]%
	\centering
	\includegraphics[width=0.5\textwidth]{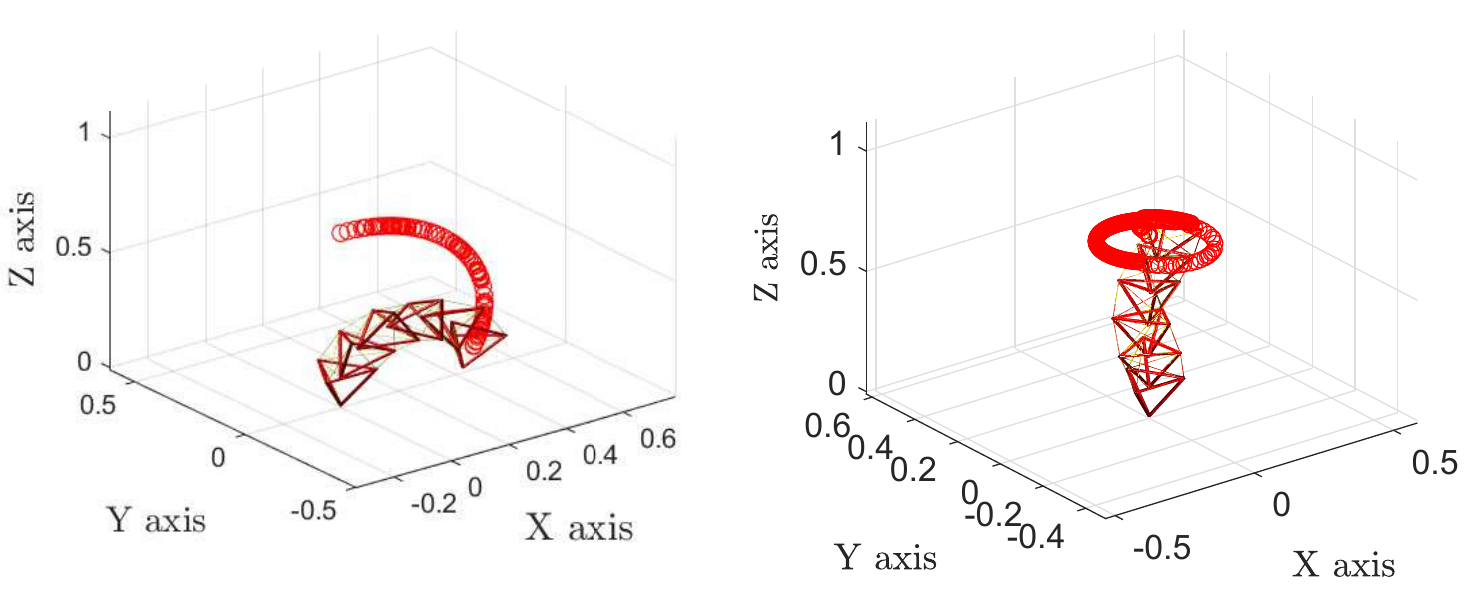}
	\caption{Simulated trajectory profile of end-effector: Bending and Axial twist configurations }
	\label{fig:wpa}
\end{figure}

\section{Results}
\begin{figure}[!htb]
	\centering
	\includegraphics[width=\linewidth, keepaspectratio]{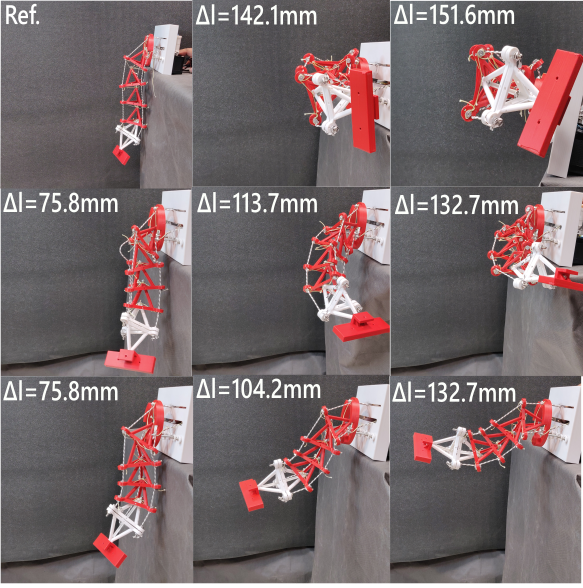}
	\caption{Sequential figures of bending generated by single force (Top), two tensile forces (Middle) and three forces (Bottom) shows the working phases of Hedra. The first configuration is the reference configuration}
	\label{fig:rsd}
\end{figure}

The simulated trajectory profile of the end-effector is shown in \ref{fig:wpa}. The sequential figures of bending  motion generated by different tensile forces and the compact configurations of the Hedra robot is shown in Fig. \ref{fig:rsd}. Experiments were conducted in two phases. Firstly, the working phases based on different type of actuation scheme of cables were investigated. Secondly, preliminary evaluation of the integrated robotic arm with gripper for potential utility in object manipulation is performed.

The performance of the integrated tensegrity robot was evaluated by employing the three types of actuation modes. The bending of the robot driven by tensile force exerted by only one cable, two cable and all three cables in tension for the minimally actuated prototype. The prototype was mounted on to the base and a simple feed-forward control structure that actuate one cable at a time was performed. A control program was developed using Arduino IDE which commands the motor driver through the IDE standard serial protocol. The serial-parallel compliant structure facilitated the possibility of simple control. By providing one tensile force to the distal end of the robot, lateral bending was observed (normal to the path of cable in tension) and the maximum bent angle was observed to be larger than 120 degrees and the change in cable length was noted. 

\begin{figure}[!htb]
	\centering
	\includegraphics[width=0.95\linewidth,trim = 2mm 2mm 2mm 2mm, clip,keepaspectratio]{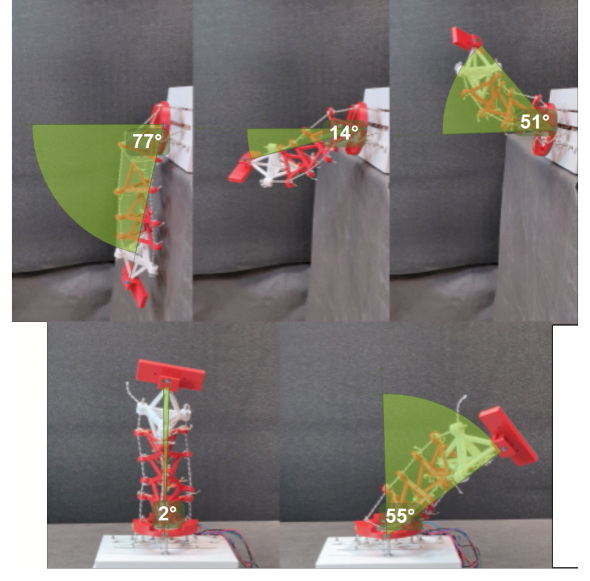} \caption{The motion of end-effector was tracked for validating a simulated trajectory and the bending angle was measured, which is shown in the green shaded area and the tracked motion is represented by the arc of the shaded area.}
	\label{fig:kin1}
\end{figure}  

The manipulator exhibits characteristics of variable stiffness by choosing the number of cables being pulled/ actuated. By pulling two opposite cables, directional stiffness can be achieved and bending happens normal to the path of cable not in tension. The maximum bent angle is lesser than the one observed in the first actuation mode. When all three cables are pulled at the same time, pure contraction happens and manipulator becomes fully stiff. The preliminary investigation proves the conceptual design of the tensegrity robot that exhibits mixed bending and contracting motion. 

\begin{figure}[th]
	\centering
	\includegraphics[width=\linewidth,keepaspectratio]{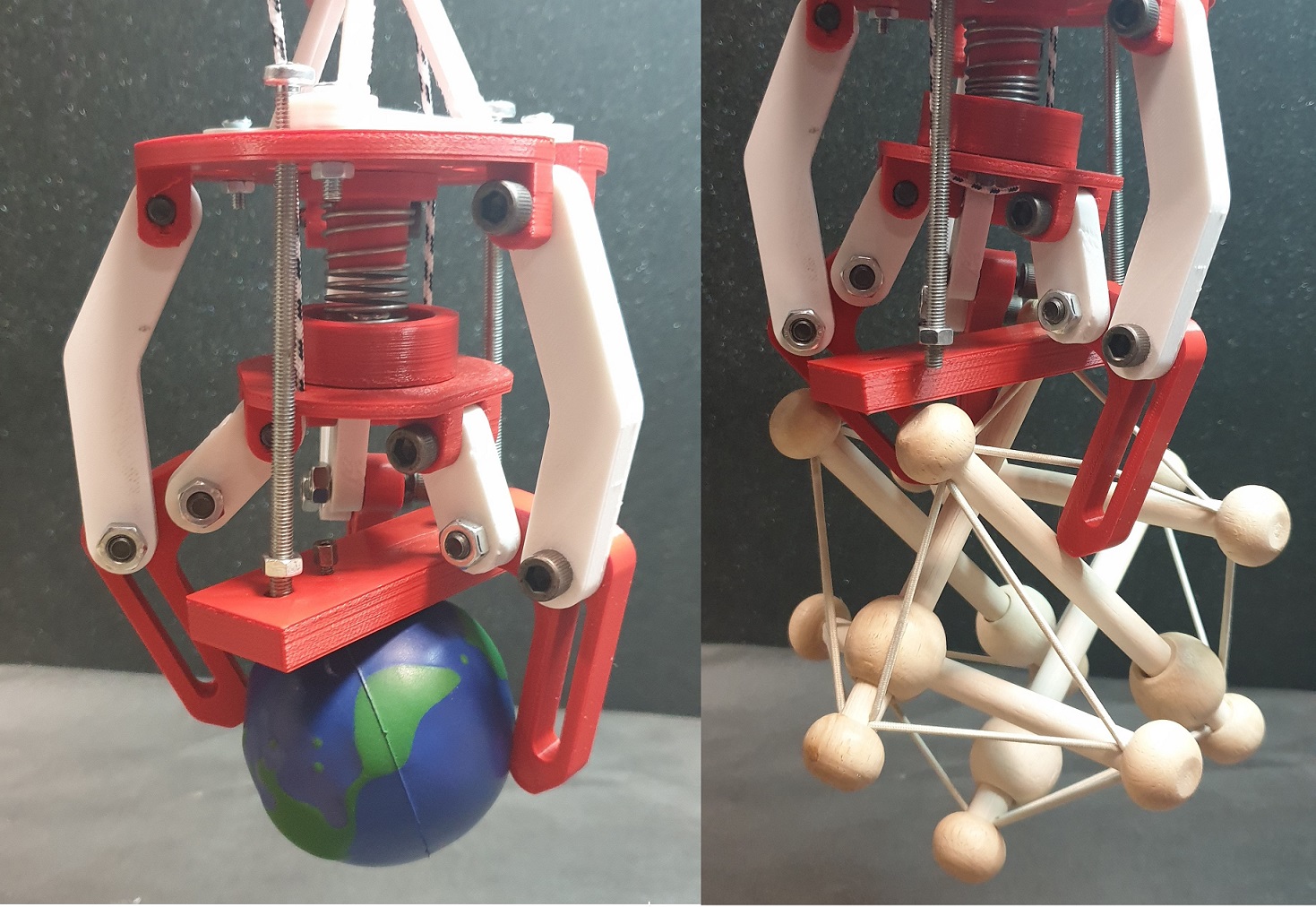}
	\caption{Left: Grasping a soft ball (weight-24 gms, material- rubber), Right: Grasping a Tensegrity toy (weight-97 gms, material- dowel rod)}
	\label{fig:rsb}
\end{figure}

The second phase of experiments were assessment of the bending angle, range of motion of the end-effector and grasp evaluation. Each motion of the manipulator was tracked for validating a simulated trajectory and the bending angle was measured. The range of motion for the top configuration shown in Fig. \ref{fig:kin1} is from -77 degrees to +51 degrees and the bending angle observed in the bottom configuration is 55 degrees. Fig. \ref{fig:rsb} shows the manipulation tasks performed by the gripper enabled Hedra. The robot was able to grasp the object placed at the platform and bend in different directions and place it back to home position. 25 trials were performed on each object, out of which 19 trials were successful for grasping and manipulating a soft ball and 23 trials were succesful for the tensegrity toy. The configuration of the stacked tensegrity manipulator is dependent on mass of the object grasped as it affects the bending range. 

\section{CONCLUSIONS}
This article provides a novel reconfigurable modular robot by integrating tensegrity-enabled compliant modules. The design process of the bio-tensegrity manipulator in three-dimensions based on the tetrahedral parallel structure linked by tensegrity joint is presented. The tensegrity joint augmented the omnidirectional compliance to the manipulator which can be safer to operate alongside humans. The horizontal saddle cables provides a high translational stiffness and supports the tensegrity joint in suspension while the active cables maintain the antagonistic rotational stiffness in the bending direction. A prototype of the novel robot with six integrated modules was fabricated and the proof-of-concept  was demonstrated showing the bending, axial twist and contraction. The manipulator was equipped with a gripper and camera for object detection and tracking. The preliminary experiments to test the manipulation capabilities were performed with a soft ball and a tensegrity toy. It was observed that the weight applied at the end effector had significant influence to the bending angles. Future work will focus on utilizing the camera system to have precise control for object detection and manipulation.

\bibliographystyle{IEEEtran}
\bibliography{hedra_main}

\end{document}